\documentclass[sigconf]{acmart}
\makeatletter
\def\@ACM@checkaffil{
}
\makeatother
 
\usepackage{multirow} 
\usepackage{amssymb}
\usepackage[ruled]{algorithm2e}
\usepackage{caption}
\usepackage{graphicx}
\usepackage{float} 
\usepackage{subfigure}
\usepackage{wrapfig}
\usepackage{hyperref}
\usepackage{makecell}

\usepackage{bbm,dsfont}
\usepackage{balance}

\AtBeginDocument{%
  }

\begin{document}


\title{ Unsupervised Visible-Infrared Person ReID by Collaborative Learning with Neighbor-Guided Label Refinement}

\author{De Cheng$^*$}
\affiliation{dcheng@xidian.edu.cn\\
\institution{Xidian University}
 }

\author{Xiaojian Huang$^*$}
\affiliation{19050300035@stu.xidian.edu.cn\\
\institution{Xidian University}
 }

\author{Nannan Wang$^{\dag}$}
\affiliation{nnwang@xidian.edu.cn\\
\institution{Xidian University}
 }

\author{Lingfeng He}
\affiliation{19169100003@stu.xidian.edu.cn\\
\institution{Xidian University} 
}

\author{Zhihui Li}
\affiliation{zhihuilics@gmail.com\\
\institution{University of New South Wales} 
}

\author{Xinbo Gao}
\affiliation{gaoxb@cqupt.edu.cn\\
\institution{Chongqing University of Posts and Telecommunications} 
}
\renewcommand{\shortauthors}{De Cheng et al.}

\thanks{$^*$Equal Contribution.}
\thanks{$^{\dag}$Corresponding author.}


\begin{abstract}
Unsupervised learning visible-infrared person re-identification (USL-VI-ReID) aims at learning modality-invariant features from unlabeled cross-modality dataset, which is crucial for practical applications in video surveillance systems.
The key to essentially address the USL-VI-ReID task is to solve the cross-modality data association problem for further heterogeneous joint learning. 
To address this issue, we propose a Dual Optimal Transport Label Assignment (DOTLA) framework to simultaneously assign the generated labels from one modality to its counterpart modality. The proposed DOTLA mechanism formulates a mutual reinforcement and efficient solution to cross-modality data association, which could effectively reduce the side-effects of some insufficient and noisy label associations. 
Besides, we further propose a cross-modality neighbor consistency guided label refinement and regularization module, to eliminate the negative effects brought by the inaccurate supervised signals, under the assumption that the prediction or label distribution of each example should be similar to its nearest neighbors’. 
Extensive experimental results on the public SYSU-MM01 and RegDB datasets demonstrate the effectiveness of the proposed method, surpassing existing state-of-the-art approach by a large margin of 7.76\% mAP on average, which even surpasses some supervised VI-ReID methods.
 
\end{abstract}


\maketitle

\section{Introduction}
Visible-infrared person re-identification (VI-ReID) aims at matching individual pedestrian images in one modality to the ones in another modality. 
Many remarkable VI-ReID works~\cite{zhang2021towards,ye2019bi,li2022counterfactual,jiang2022cross,zhang2022modality,liu2022learning,zhang2022fmcnet} can effectively address the applicability limitation of single-modality person ReID in practical surveillance systems, where conventional ReID can not work well for images captured under the poor lighting conditions, especially under the night conditions. Therefore, VI-ReID has attracted increasing attention due to its importance in video surveillance applications. 

However, training VI-ReID models in a supervised manner requires 
a substantial amount of cross-modality identity annotations, making it expensive, labor-intensive, and limiting its practical application in real-world scenarios. 
Thus, the development of an effective unsupervised method for VI-ReID using unlabeled data is highly appealing and significant, not only in the academic sector but also in industrial fields.

To this end, this paper aims at learning modality-invariant features from the unlabeled visible-infrared dataset, and the key challenge is how to effectively generate reliable visible-infrared positive labels.
Existing state-of-the-art unsupervised single-modality person ReID methods usually leverage the pseudo labels obtained from clustering results to train the deep model. The training scheme of these methods usually alternates between clustering to generate pseudo labels and fine-tuning the deep model with these pseudo labels in a supervised manner. Such pseudo-label-based unsupervised learning pipeline has yielded promising results on the single-modality unsupervised ReID task. 
However, for USL-VI-ReID, the cross-modality discrepancy between the visible and infrared images is very large, which makes it very difficult to generate visible-infrared positive labels for cross-modality training. 

To address the challenges in USL-VI-ReID, we propose a two-stream collaborative learning framework followed by the neighbor consistency guided label refinement and regularization modules, which gradually reduces the modality discrepancy and further benefits the cross-modality positive labels generation. Specifically, we first leverage existing advanced single-modality unsupervised learning technique to initialize reliable pseudo labels for these two modality images, respectively. Then we train the modality-shared network with these pseudo labels from each modality, aiming to learn modality-invariant features. 
In the following, our goal is to solve the cross-modality label association problem for further heterogeneous joint learning, to essentially address the USL-VI-ReID problem.

Specifically, we propose a Dual Optimal Transport Label Assignment (DOTLA) strategy to simultaneously assign the generated labels from one modality to another modality images. Such DOTLA strategy establishes an explicit connection between cross-modality data and ensures that the optimization explicitly concentrates on the modality irrelevant perspective. Compared with the unidirectional assignment, the dual label assignment strategy formulates a mutual reinforcement and efficient solution to cross-modality data association. Besides, the DOTLA strategy takes into account both of the explicit sample-to-cluster and implicit sample-to-sample association, which could effectively reduce the side-effects brought by some insufficient and noisy label associations. Afterwards, we could build the modality-agnostic objective function on top of these label assignment results, to learn modality-invariant and discriminative feature representations. Finally, the training process alternates between the cross-modality label assignments and network optimization, which gradually reduces the large modality discrepancy and allows us to find truly matched visible-infrared samples.

However, there inevitably contains some wrong label associations among the generated pseudo labels. To eliminate the negative effects brought by the inaccurate supervisory signals, we further propose a neighbor-consistency guided label refinement module. We conduct label refinement under the assumption that the prediction of each example should be similar to its nearest
neighbours’~\cite{iscen2022learning}.
It seeks to transfer labels from its neighbouring instances, and encourages each example to have similar prediction to its neighbours’. 
Specifically, we have designed a criterion to first identify whether an instance is noisy or not using the neighbors from its counterpart modality, and the cross-modality label refinement is only conducted on the noisy labels. Furthermore, the cross-modality neighbor consistency regularization has also been proposed to align the prediction distribution of each instance and its cross-modality neighbors if they are close in the feature space from a local perspective, so as to enhance the
neighbour consistency assumption.


The main contributions can be summarized as follows:
\begin{itemize}
    \item  We propose a dual optimal transport label assignment framework to essentially address the cross-modality data association problem, which simultaneously assigns the generated labels from one modality to another modality images. Such dual label assignment strategy formulates a mutual reinforcement and efficient solution to effectively reduce the side-effects of insufficient and noisy label associations.

    \item To eliminate the negative effects brought by the inaccurate supervisory signals, we further propose the neighbor consistency guided label refinement and regularization modules, under the assumption that the predicted label distribution of each example should be similar to its nearest neighbors'.

    \item Extensive experimental results on the SYSU-MM01 and RegDB datasets demonstrate the effectiveness of the proposed method, surpassing existing state-of-the-art approach by a large margin of 7.76\% mAP on average, which even surpasses some supervised VI-ReID methods. 
    
\end{itemize}

\section{RELATED WORK}
\subsection{Visible-Infrared Person ReID}
Visible-infrared person re-identification (VI-ReID) aims to match pedestrian images from different modalities. The main challenge is to reduce the modality gap. Recently, some works \cite{wang2020cross,choi2020hi,wang2019rgb,wang2019learning} focus on constructing effective transformation between two modalities that preserve identity information to align visible and infrared modalities. 
These methods mainly use generative models for information transformation. 
Another branch of works adopts an intermediate modality to assist with modality gap reduction. Some representative works \cite{wei2021syncretic, li2020infrared}
design a lightweight network to generate the intermediate modality, while Ye \textit{et al.} \cite{ye2021channel} proposes a channel augmentation technique to directly generate an intermediate modality from visible modality. Additionally, many recent works~\cite{ye2020dynamic,lu2020cross, wu2021discover,zhang2022modality} explore to construct modality-shared and modality-specific feature spaces for VI-ReID. However, the methods mentioned above require large-scale cross-modality labeled datasets, which are expensive, time-consuming and laborious to be obtained. 


\begin{figure*}
  \centering
  \includegraphics[scale=0.93]{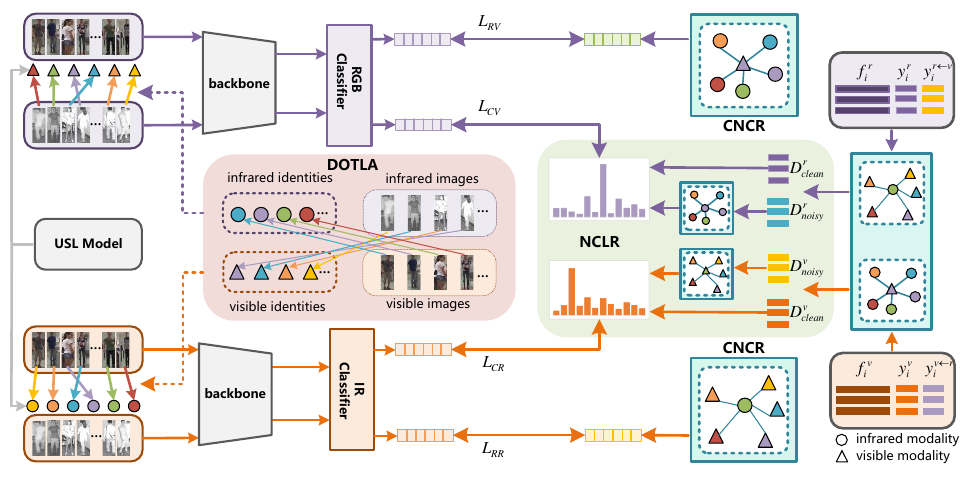}\\
  \caption{Overall framework of our proposed method. The framework mainly contains three components: Dual Optimal Transport Label Assignment(DOTLA), Neighbor Consistency guided Label Refinement(NCLR), and Cross-modality Neighbor Consistency Regularization(CNCR). Circle and triangle represent instances from infrared modality and visible modality. Different colors stand for different identities.  }\label{framework}
\end{figure*}

\subsection{Unsupervised Single-Modality Person ReID}
The traditional unsupervised person ReID methods can be roughly divided into two categories: fully unsupervised learning (USL) methods and unsupervised domain adaptation (UDA) methods, depending on whether or not the labeled source domain datasets are utilized for model training. 
For UDA, some works~\cite{wei2018person,zhong2018generalizing,chen2019instance,zou2020joint,chen2021joint} leverage generative methods to transfer knowledge from the source domain to the target domain, while others~\cite{ge2020mutual,zhai2020multiple,zhai2020ad,dai2021dual} follow a two-stage process: training model on source domain in a supervised manner, and then iteratively generating pseudo-labels and fine-tuning the model with the unlabeled target domain dataset. 
In recent years, many USL methods~\cite{dai2022cluster,wang2021camera} usually leverage pseudo labels generated by clustering method to optimize the model. 
To further improve the quality of the pseudo labels, 
ISE~\cite{zhang2022implicit} designs an implicit sample extension module to improve the quality of clusters. PPLR~\cite{cho2022part} utilizes part features to refine pseudo labels. 
However, above-mentioned methods only address the unsupervised single-modality ReID task. When dealing with the USL-VI-ReID task, the pseudo labels generated by directly using the clustering method become unreliable because of the large modality discrepancy.

\subsection{Unsupervised Visible-Infrared Person ReID}
The key to USL-VI-ReID task is to establish data association between modalities without identity annotations.
H2H~\cite{liang2021homogeneous} is the first to attempt to solve this task by proposing a two-stage learning framework. It firstly obtains reliable pseudo labels through a clustering method and then aligns two modalities by cross-modality pair-wise instances selection. 
Wang \textit{et al.} \cite{wang2022optimal} formulates the visible-infrared data association problem as an optimal transport problem to assign visible-modality labels to the infrared images. 
ADCA\cite{yang2022augmented} proposes a dual-contrastive learning framework that utilizes the augmented intermediate images and cross-modality memory aggregation module to help reduce the modality gap. 

Compared to the above methods, we focus on handling noisy cross-modality label assignments. We propose a dual optimal transport label assignment strategy and a neighbor consistency guided label refinement  module to generate soft labels for the whole dataset and then design a cross-modality neighbor consistency regularization to further eliminate the side-effects brought by noisy labels. 

\section{PROPOSED METHOD}
\subsection{Problem Formulation and Overview}
Let $\mathcal{X}=\{\mathcal{V}, \mathcal{R}\}$ denote the unsupervised learning visible-infrared ReID (USL-VI-ReID) dataset, $\mathcal{V}=\{\mathbf{x}_i^v \}_{i=1}^{N_v}$ and $\mathcal{R}=\left\{\mathbf{x}_i^r\right\}_{i=1}^{N_r}$ represent the visible and infrared subset with $N_v$ and $N_r$ samples, respectively. The USL-VI-ReID dataset does not contain any identity label for each image. For USL-VI-ReID task, the goal is to train a deep neural network $f_{\theta}$ with the unlabeled dataset to project the input image $\mathbf{x}_i$ into the modality-shared feature space $\mathcal{F}_{\theta}$, then we can match the images of the same person from different two modalities in the feature space $\mathcal{F}_{\theta}$.


In this paper, we propose an efficient collaborative learning framework with cross-modality neighbor consistency guided label refinement for USL-VI-ReID, as shown in Figure~\ref{framework}. The proposed method has three components including the collaborative learning mechanism by Dual Optimal Transport Label Assignment (DOTLA) mechanism, the cross-modality Neighbor Consistency guided Label Refinement (NCLR) module, and the Cross-modality Neighbor Consistency Regularization (CNCR) module. 
These three components aim to address the key challenge for USL-VI-ReID, which is how to effectively generate reliable cross-modality visible-infrared positive labels and further learn modality-invariant features.


Specifically, by virtue of the remarkable progress in unsupervised single-modality ReID, we first generate reliable pseudo labels for both the visible and infrared subsets independently, using the state-of-the-art USL-ReID method Cluster-Contrast~\cite{dai2022cluster}. 
Afterwards, the cross-modality discrepancy can be gradually reduced through the following techniques. 
Given these reliable single-modality pseudo labels, we then train the shared backbone network with two modality-specific classifiers independently, to learn modality-invariant features by projecting these two modality images $\mathbf{x}_i$ into one common feature space $\mathcal{F}_{\theta}$. 

Most importantly, to align the pseudo labels from two different modalities, the cross-modality label matching mechanism DOTLA is proposed to transport the infrared/visible images to the generated visible/infrared pseudo label space,
enabling further joint heterogeneous learning. 
This bi-directional label assignment mechanism formulates a mutual reinforcement solution to cross-modality data association, which could reduce the side-effects of some insufficient and noisy label associations to a certain extent. Furthermore, we leverage the NCLR and the CNCR module to facilitate the global label assignment by DOTLA, and thus eliminating the negative effects brought by inaccurate supervised signals to some extent. 
Finally, the proposed method could gradually learn better modality-invariant features, through mining more and more accurate  association knowledge between modalities.

\subsection{Dual Optimal Transport Label Assignment}
Let $\left\{y_i^v\right\}_{i=1}^{N_v}$ and $\left\{y_i^r\right\}_{i=1}^{N_r}$ define the initial pseudo labels for the visible and infrared subsets $\mathcal{V}=\{\mathbf{x}_i^v \}_{i=1}^{N_v}$ and $\mathcal{R}=\left\{\mathbf{x}_i^r\right\}_{i=1}^{N_r}$, respectively. They are obtained by conducting one representative unsupervised single-modality person ReID method (ClusterContrast~\cite{dai2022cluster}) on each modality subset, respectively. Note that these pseudo labels from two different modalities are not aligned. 

At first, we design two modality-specific classifiers (i.e., $\phi_v$ for visible and $\phi_r$ for infrared ), on top of the shared backbone network $\mathcal{F}_{\theta}$. Given the visible and infrared images (i.e., $\mathbf{x}_i^v$ and $\mathbf{x}_i^r$), their corresponding features and classifier predictions can be represented as:  $\mathbf{f}_i^v = f_{\theta}(\mathbf{x}_i^v)$, $\mathbf{f}_i^r = f_{\theta}(\mathbf{x}_i^r)$ and $p_i^v = \phi_v(\mathbf{x}_i^v;\theta)$, $p_i^r = \phi_r(\mathbf{x}_i^r;\theta)$, respectively. Initially, we optimize the network parameters $\{\theta,\phi_v,\phi_r\}$ with the standard triplet loss and cross-entropy loss using the previously obtained unaligned pseudo labels $\left\{y_i^v\right\}_{i=1}^{N_v}$ and $\left\{y_i^r\right\}_{i=1}^{N_r}$, as follows,   
 
\begin{flalign} 
&\mathcal{L}_{v}^{reid}(\theta,\phi_v) = \mathcal{L}_{v}^{tri}(\theta) + \mathcal{L}_{v}^{ce}(\theta, \phi_v)\label{visibleLoss}\\
&\mathcal{L}_{r}^{reid}(\theta,\phi_r) = \mathcal{L}_{r}^{tri}(\theta) + \mathcal{L}_{r}^{ce}(\theta,\phi_r)\label{infraredLoss}.
\end{flalign}
In Eq.~\ref{visibleLoss}, $\mathcal{L}_{v}^{tri}(\theta)$ and $\mathcal{L}_{v}^{ce}(\theta,\phi_v)$ denote the standard triplet loss and cross-entropy loss for the visible modality, and their combination (denoted as $\mathcal{L}_{v}^{reid}(\theta,\phi_v)$) can be regarded as the conventional ReID loss on single-modality dataset. Note that $\mathcal{L}_{v}^{reid}(\theta,\phi_v)$ just works on the visible modality subset with the given pseudo labels  $\left\{y_i^v\right\}_{i=1}^{N_v}$. Correspondingly, in Eq.~\ref{infraredLoss}, the ReID loss on infrared modality $\mathcal{L}_{r}^{reid}(\theta,\phi_r)$, which is a combination of triplet loss $\mathcal{L}_{r}^{tri}(\theta)$ and cross-entropy loss $\mathcal{L}_{r}^{ce}(\theta,\phi_r)$, just works on the infrared modality subset with the given pseudo labels $\left\{y_i^r\right\}_{i=1}^{N_r}$.
Jointly optimizing the loss functions $\mathcal{L}_{v}^{reid}(\theta,\phi_v)$ and $\mathcal{L}_{r}^{reid}(\theta,\phi_r)$ with independent pseudo label subsets initially, can  
encourage the shared backbone network to learn modality-invariant features, thus reducing the modality gap preliminarily.


In the following, we focus on the key challenge of USL-VI-ReID problem: cross-modality data association. To solve such an issue, we propose a Dual Optimal Transport Label Assignment(DOTLA) module for collaborative learning. It simultaneously matches the infrared images $\mathcal{R}=\left\{\mathbf{x}_i^r\right\}_{i=1}^{N_r}$ to the visible label space $\mathcal{Y}^v$ and obtains $\left\{y_i^{r \gets v}\right\}_{i=1}^{N_r}$, as well as matching the visible images $\mathcal{V}=\left\{\mathbf{x}_i^v\right\}_{i=1}^{N_v}$ to the infrared label space $\mathcal{Y}^r$ and obtaining $\left\{y_i^{v \gets r}\right\}_{i=1}^{N_v}$. Different from the unidirectional label assignment~\cite{wang2022optimal}, the DOTLA strategy formulates a mutual reinforcement solution to cross-modality data association, which could effectively reduce the side-effects of some noisy label associations. 

The proposed \textbf{collaborative learning} framework consists of two branches: the \textbf{visible-ReID branch} and \textbf{infrared-ReID branch}. Both branches share the same backbone network parameters $\theta$, while with independent classifier head parameters, i.e., $\phi_v$ and $\phi_r$. In the following collaborative training procedure, both branches use the whole cross-modality training dataset but with different assigned label sets.  

For the \textbf{visible-ReID branch}, we train the network with datasets $\mathcal{V}=\{\mathbf{x}_i^v \}_{i=1}^{N_v}$ and $\mathcal{R}=\{\mathbf{x}_i^r \}_{i=1}^{N_r}$, and their corresponding labels are $\left\{y_i^v\right\}_{i=1}^{N_v}$ and  $\left\{y_i^{r \gets v}\right\}_{i=1}^{N_r}$, respectively.
$y_i^{r \gets v}$ is the assigned label for example $\mathbf{x}_i^r$ from the visible label space $\mathcal{Y}^v$ by DOTLA module. We formulate the cross-modality label assignment task as an optimal transport problem, and the objective function for assigning labels to infrared images from the visible label space can be described as:

\begin{equation}
    \begin{aligned}
   &\min_{\mathbf{Q}^r} <\mathbf{Q}^r, -\log(\mathbf{P}^r)> + \frac{1}{\lambda} KL(\mathbf{Q}^r||\boldsymbol{\alpha} \boldsymbol{\beta}^{\top}), \\ 
   \vspace{1ex}
    &s.t. 
    \begin{cases}
    \vspace{1ex}\mathbf{Q}^r \mathds{1} = \boldsymbol{\alpha}, \boldsymbol{\alpha} = \mathds{1} \cdot \frac{1}{N_r}, \\ 
    (\mathbf{Q}^r)^{\top} \mathds{1} = \boldsymbol{\beta}, \boldsymbol{\beta} = \mathds{1} \cdot \frac{1}{N_p^v}, 
    \end{cases}
    \label{eq:DOTLA}
    \end{aligned}
\end{equation}
where $\mathbf{Q}^r \in \mathbb{R}^{N_r \times N_p^v}$ is the optimal plan for pseudo label assignment from visible label space to the infrared images, $\mathbf{P}^r \in \mathbb{R}^{N_r \times N_p^v}$ represents the predictions of the visible classifier for infrared images, and $N_p^v$ is the number of visible identities, $<\cdot,\cdot>$ denotes the Frobenius dot-product, $KL(\cdot||\cdot)$ denotes the KL-Divergence, and $\mathbbm{1}$ is an all in 1 vector. We can clearly see that the transport problem is also a trade-off between prediction and the smooth assignment. $\boldsymbol{\alpha} \in \mathbb{R}^{N_r \times 1}$ and $\boldsymbol{\beta} \in \mathbb{R}^{N_p^v \times 1}$ represent the constraints that an infrared image is assigned to exactly one label from the visible label space and that $N_r$ infrared images should be split into $N_p^v$ classes equally. 

We use the Sinkhorn-Knopp algorithm~\cite{cuturi2013sinkhorn} to solve this objective in Eq.~\ref{eq:DOTLA}, and obtain the optimal pseudo label assignment plan $\mathbf{Q}^r \in \mathbb{R}^{N_r \times N_p^v}$. Finally, the label assignment results for the infrared images can be obtained as $\left\{y_i^{r \gets v}\right\}_{i=1}^{N_r}$ according to $\mathbf{Q}^r$. Then, we minimize the following objective function $\mathcal{L}_{CV}$ for the visible-ReID branch in the collaborative learning framework, to optimize the deep model,  
\begin{align}
\mathcal{L}_{CV} = \frac{1}{N_{r}} \sum_{i=1}^{N_{r}} \mathcal{L}_{ce}(\phi_v(\mathbf{x}_i^{r};\theta), y_i^{r\gets v}) + \mathcal{L}_{v}^{reid}(\theta,\phi_v).
\label{visibleBranch}
\end{align}

Accordingly, for the \textbf{infrared-ReID branch}, we also assign labels to the visible images from the infrared label space, by optimizing the optimal transport objective as shown in Eq.~\ref{eq:DOTLA}. We train the network on both visible and infrared datasets $\mathcal{V}$ and $\mathcal{R}$, with the corresponding labels as $\left\{y_i^{v \gets r}\right\}_{i=1}^{N_v}$ and $\left\{y_i^r\right\}_{i=1}^{N_r}$, respectively. Simultaneously, we minimize the following objective function $\mathcal{L}_{CR}$ for the infrared-ReID branch in the collaborative learning framework, as follows,
\begin{align}
\mathcal{L}_{CR} = \frac{1}{N_{v}} \sum_{i=1}^{N_{v}} \mathcal{L}_{ce}(\phi_r(\mathbf{x}_i^{v};\theta), y_i^{v\gets r}) + \mathcal{L}_{r}^{reid}(\theta,\phi_r).
\label{infraredBranch}
\end{align}

Jointly optimizing Eq.~\ref{visibleBranch} and Eq.~\ref{infraredBranch} encourages learning modality-shared backbone network parameterized by $\theta$. 
Experiments demonstrate that such a collaborative learning mechanism helps to effectively reduce the label assignment noise between modalities.

\subsection{Neighbor Consistency Guided Label Refinement}\label{sec:labelrefinement}
Although the DOTLA module has established explicit data association between cross-modality data, there inevitably contains some wrong label associations, and the assigned pseudo labels could be noisy, which will negatively affect the model optimization. Therefore, we propose the label refinement module guided by the cross-modality neighbor consistency, under the assumption that \emph{the label prediction of each example should be similar to its nearest neighbors'}, where the nearest neighbors could be obtained within the same-/cross-modality feature space.  The label refinement module could be applied to both the visible and infrared ReID branches in collaborative learning. In the following, we just take the visible-ReID branch for illustration. 


Specifically, label refinement requires two steps: 1) Noisy label identification; 2) Noisy label refinement. Firstly, we design a criterion to identify whether an instance is noisy or not, which measures the similarity between the label of the instance and its neighbors'. Given an instance with the assigned label from cross-modality label space $(\textbf{x}_i^{r}, y_i^{r \gets v})$, the inconsistency score of the assigned label $y_i^{r \gets v}$ for the instance $\textbf{x}^r_i$ with its neighbours can be defined as $s(\textbf{x}^r_i)$. It can be calculated by the KL divergence between $y_i^{r \gets v}$ and the mean prediction of its neighbors' in the visible-modality space, as follows,
\begin{flalign}\label{eq:k}
    s(\mathbf{x}_i^{r}) = KL(y_i^{r \gets v}||\frac{1}{k}\sum_{\textbf{x}_j^{v} \in  \mathcal{N}_k^v }y_j^{v}),
\end{flalign}
where $\mathcal{N}_k^v=\mathcal{N}(\textbf{x}_i^{r},\mathcal{V}, k)$ denotes the $k$-nearest neighbors of $\textbf{x}_i^r$ in the visible-modality space $\mathcal{V}$, and $y_j^{v}$ is the original single-modality label for $\textbf{x}_i^v$. Note that, the $k$-nearest neighbors are computed by the $L_2$ distance in the feature space. The smaller of $s(\textbf{x}_i^{r})$ means the higher similarity between the assigned label and its neighbors' predictions, which usually reveals that the assigned label  $y_i^{r \gets v}$ is clean, vice verse.


Afterwards, the infrared-modality data can be divided into the clean subset and noisy subset according to the inconsistency score $s(x_i^{r})$ on each instance $x_i^{r}$, which can be expressed as follows,
\begin{flalign}\label{eq:tau}
    &\mathcal{D}_{clean}^r = \{\textbf{x}_i^r|s(\textbf{x}_i^r) \le \tau, \textbf{x}_i^r \in \mathcal{R}\}, \\
    &\mathcal{D}_{noisy}^r = \{\textbf{x}_i^r|s(\textbf{x}_i^r) \textgreater \tau, \textbf{x}_i^r \in \mathcal{R} \},   
\end{flalign}
where $\tau$ is a threshold parameter to distinguish the assigned label $y_i^{r\gets v}$ of instance $\textbf{x}_i^r$ to be noisy or clean. Naturally, only the assigned label  $y_i^{r\gets v}$ for the instance in the noisy subset $\textbf{x}_i^r \in \mathcal{D}_{noisy}^r$ should be refined to avoid the noisy learning. 

Specifically, we propose to use the clean neighbors $\hat{\mathcal{N}}^{r}_k$ for each instance $\textbf{x}_i^r \in \mathcal{D}_{noisy}^r$ in the single-modality space $\mathcal{R}$ to refine its assigned noisy label $y_i^{r\gets v}$. The clean neighbor set for $\textbf{x}_i^r$ can be designed as follows,
\begin{flalign}
    \hat{\mathcal{N}}^{r}_k = \{\textbf{x}_j^r|\textbf{x}_j^r \in \mathcal{N}(\textbf{x}_i^r;\mathcal{R};k) \cap \mathcal{D}_{clean}^r\},
\end{flalign}
where $\mathcal{N}(\textbf{x}_i^r;\mathcal{R};k)$ denotes the $k$-nearest neighbors of $\textbf{x}_i^r$ in the infrared-modality space $\mathcal{R}$, and the neighbors are also constrained in the clean subset $\mathcal{D}_{clean}^r$. 
Then we obtain the refined label $\hat{y}_i^{r\gets v}$ for $\textbf{x}_i^r \in \mathcal{D}_{noisy}^r$ as follows:
\begin{flalign}\label{gamma}
    \hat{y}_i^{r\gets v} = \begin{cases}
    \vspace{1ex}
    (1-\gamma) \cdot  y_i^{r \gets v} + \gamma \cdot \frac{1}{k} \sum_{\textbf{x}_j^r \in \hat{\mathcal{N}}_k^r} y_j^{r\gets v}, &\textbf{x}_i^r \in \mathcal{D}_{noisy}^r \\ 
    \vspace{1ex}
    y_i^{r \gets v}, & \textbf{x}_i^r \in \mathcal{D}_{clean}^r \end{cases}
\end{flalign}\
where $\gamma$ is a hyper-parameter that controls the degree of refinement for the assigned noisy labels, and $y_i^{r \gets v}$ is the assigned label from the visible label space $\mathcal{Y}^v$ for instance $\textbf{x}_i^r$. 

Following the above label refinement strategy, we can obtain the refined cross-modality assigned labels  $\{\hat{y}_i^{r\gets v}\}_{i=1}^{N_r}$ and  $\{\hat{y}_i^{v \gets r}\}_{i=1}^{N_v}$ for the visible-ReID branch and infrared-ReID branch, respectively. Therefore, Eq.~\ref{visibleBranch} and Eq.~\ref{infraredBranch} can be re-written as follows:
\begin{align}
&\mathcal{L}_{C\hat{V}} = \frac{1}{N_{r}} \sum_{i=1}^{N_{r}} \mathcal{L}_{ce}(\phi_v(\textbf{x}_i^{r};\theta), \hat{y}_i^{r\gets v}) + \mathcal{L}_{v}^{reid}(\theta,\phi_v),\\
&\mathcal{L}_{C\hat{R}} = \frac{1}{N_{v}} \sum_{i=1}^{N_{v}} \mathcal{L}_{ce}(\phi_r(\textbf{x}_i^{v};\theta), \hat{y}_i^{v\gets r}) + \mathcal{L}_{r}^{reid}(\theta,\phi_r).
\end{align}

\subsection{Cross-modality Neighbor Consistency Regularization}
Besides the neighbor consistency based label refinement as described in section~\ref{sec:labelrefinement}, we also propose the Cross-modality Neighbor Consistency Regularization (CNCR) module, to further eliminate the negative effects brought by the assigned noisy labels for model training. The CNCR module aligns the prediction distribution between each instance and its cross-modality neighbors if they are close in the feature space, so as to enhance the neighbor consistency assumption. It can be applied to both the visible-ReID and infrared-ReID branches.

Specifically, the CNCR module encourages the prediction of each instance to be consistent with its cross-modality neighbors', and it can be formulated through the Kullback–Leibler divergence $KL(\cdot || \cdot)$ for both branches respectively, as follows:
\begin{align}
    &\mathcal{L}_{RV}= \frac{1}{N_v} \sum_{i=1}^{N_v} KL(\phi_v(\textbf{x}_i^{v};\theta)||\frac{1}{k} \sum_{\textbf{x}_j^r \in \mathcal{N}_k^{vr}} \phi_v(x_j^{r};\theta)), \\
    &\mathcal{L}_{RR} = \frac{1}{N_r} \sum_{i=1}^{N_r} KL(\phi_r(\textbf{x}_i^{r};\theta)||\frac{1}{k} \sum_{\textbf{x}_j^v \in \mathcal{N}_k^{rv}} \phi_r(\textbf{x}_j^{v};\theta)),
\end{align}
where $\mathcal{N}_k^{vr}=\mathcal{N}(\textbf{x}_i^v, \mathcal{R}, k)$ denotes the $k$-nearest neighbors of $\textbf{x}_i^v$ in the infrared-modality space $\mathcal{R}$, and $\mathcal{N}_k^{rv}=\mathcal{N}(\textbf{x}_i^r, \mathcal{V}, k)$ denotes the $k$-nearest neighbors of $\textbf{x}_i^r$ in the visible-modality space $\mathcal{V}$. Clearly, we can see that $\mathcal{L}_{RV}$ and $\mathcal{L}_{RR}$ show the cross-modality regularization. Note that, the $k$-nearest neighbors here are searched within each batch for efficient model training. To simplify, we re-write the two-branch cross-modality neighbor consistency regularization as follows,  
\begin{flalign}
    \mathcal{L}_{R}= \mathcal{L}_{RV}+ \mathcal{L}_{RR}.
\end{flalign}

\subsection{Optimization}
In the beginning, the model is optimized with loss functions defined below:
\begin{align}
    \mathcal{L} = \mathcal{L}_{v}^{reid} + \mathcal{L}_{r}^{reid}.
\label{equ:baseline}
\end{align}
After training several epochs to obtain effective modality-shared features, we will leverage the DOTLA module for cross-modality label association, following the neighbor consistency guided label refinement and regularization, to try to establish a more accurate association between two modalities.
Finally, we optimize the deep model with loss function as follows:
\begin{flalign}\label{alpha}
    \mathcal{L} = \mathcal{L}_{C\hat{V}} + \mathcal{L}_{C\hat{R}} +\alpha \mathcal{L}_{R},
\end{flalign}
where $\alpha$ is a hyper-parameter to balance the ReID loss and the cross-modality neighbour consistency regularization term $\mathcal{L}_{R}$. 

\begin{table*}
\small
  \centering
  \vspace{-0.3cm}
\caption{Comparison with the state-of-the-art methods on SYSU-MM01 dataset. It contains supervised VI-ReID (SVI-ReID), unsupervised single-modality ReID (USL-ReID), and unsupervised VI-ReID (USL-VI-ReID) methods. Rank-k accuracy(\%), mAP (\%) and mINP (\%) are reported.}
 \label{table 1}

	\begin{tabular}{c | c c | c c c c c | c c c c c}
		\toprule

		&\multicolumn{2}{c|}{SYSU-MM01 Settings} & \multicolumn{5}{c|}{All Search} & \multicolumn{5}{c}{Indoor Search} \\

		&Methods&Venue&r1(\%)&r10(\%)&r20(\%) &mAP(\%)&mINP(\%)&r1(\%)&r10(\%)&r20(\%)&mAP(\%)&mINP(\%)\\
		\hline 
		\multirow{5}{*}{SVI-ReID}
            &X-Modal~\cite{li2020infrared}&AAAI-20&49.9&89.8&96.0&50.7&-&-&-&-&-&-\\
		&Hi-CMD~\cite{choi2020hi}&CVPR-20&34.9&77.6&-&35.9&-&-&-&-&-&-\\
	
		&AGW~\cite{ye2021deep}&TPAMI-21&47.50&84.39&92.14&47.65&35.30&54.17&91.14&95.98&62.97&59.23\\
		&CA~\cite{ye2021channel}&ICCV-21&69.88&95.71&98.46&66.89&53.61&76.26&97.88&99.49&80.37&76.79\\
		&FMCNet~\cite{zhang2022fmcnet}&CVPR-22&66.34&-&-&62.51&-&68.15&-&-&74.09&-\\
		\hline 
		\multirow{6}{*}{USL-ReID}&SPCL~\cite{ge2020self}&NIPS-20&18.37&54.08&69.02&19.39&10.99&26.83&68.31&83.24&36.42&33.05\\
		&MMT~\cite{ge2020mutual}&ICLR-20&21.47&59.65&73.29&21.53&11.50&22.79&63.18&79.04&31.50&27.66\\
		&Cluster-Contrast~\cite{dai2022cluster}&arXiv-21&20.16&59.27&72.5&22.00&12.97&23.33&68.13&82.66&34.01&30.88\\
		&ICE~\cite{chen2021ice}&ICCV-21&20.54&57.5&70.89&20.39&10.24&29.81&69.41&82.66&38.35&34.32\\
            &PPLR~\cite{cho2022part} &CVPR-22&12.58&47.43&62.69&12.78&4.85&13.65&52.66&70.28&22.19&18.35
\\
            & ISE \cite{zhang2022implicit} & CVPR-22 & 20.01 & 57.45 & 72.50 & 18.93 & 8.54 & 14.22 & 58.33 & 75.32 & 24.62 & 21.74 \\
            \hline 
		\multirow{3}{*}{USL-VI-ReID}&H2H~\cite{liang2021homogeneous}&TIP-21&30.15&65.92&77.32&29.40&-&-&-&-&-&-\\
		&OTLA~\cite{wang2022optimal}&ECCV-22&29.98&71.79&83.85&27.13&-&29.8&-&-&38.8&-\\
		&ADCA~\cite{yang2022augmented}&MM-22&\underline{45.51}&\underline{85.29}&\underline{93.16}&\underline{42.73}&\underline{28.29}&\underline{50.60}&\underline{89.66}&\underline{96.15}&\underline{59.11}&\underline{55.17}\\
        \hline 
		&Ours&-&\textbf{50.36}&\textbf{89.02}&\textbf{95.92}&\textbf{47.36}&\textbf{32.40}&\textbf{53.47}&\textbf{92.24}&\textbf{97.84}&\textbf{61.73}&\textbf{57.35} \\
            \bottomrule
	\end{tabular}

\end{table*}

\section{EXPERIMENTS}

\subsection{Datasets and Evaluation Protocol}
We evaluate our proposed method on two public VI-ReID datasets: SYSU-MM01 and RegDB. 
Following the settings in \cite{wu2017rgb, ye2019bi}, the Cumulative Matching Characteristic (CMC) and mean Average Precision (mAP) are used as evaluation metrics. Additionally, we report the mean Inverse
Negative Penalty (mINP) \cite{ye2021deep} metric to measure the cost of finding all the correct matches.\\
\textbf{SYSU-MM01} is a large public VI-ReID dataset sampled from 4 visible cameras and 2 near-infrared cameras. The training set contains 22,258 visible images and 11,909 infrared images of 395 identities, while the test set contains 3,803 infrared images and 301 visible images
of 96 identities. We evaluate our method on two testing modes: all-search mode and indoor-search mode. For the all-search mode, the gallery set has images from all visible cameras, while for the indoor-search mode, the gallery set contains images only from the indoor visible cameras.\\
\textbf{RegDB} is captured by one visible and one
far-infrared (thermal) cameras. It consists of 4,120 images from 412 identities, with 206 identities selected for training and the remaining 206 identities for testing. Each identity has 10 visible images and 10 thermal images.
The RegDB dataset has two test modes: visible-to-infrared mode and infrared-to-visible mode. We evaluate our model under both modes.

\subsection{Implementation Details}
The proposed method is implemented on Pytorch platform. The AGW~\cite{ye2021deep} network is adopted as our backbone, which is pre-trained on the labeled Market-1501 dataset. The settings in the pre-training process follow CA~\cite{ye2021channel}. The batch size is set to 128. Each branch contains 8 person identities, and each identity contains 4 visible instances and 4 infrared instances.

All images are resized to 256$\times$128. During training, the images are augmented by random cropping, flipping, and erasing. In the first stage, we use the SGD optimizer for training with an initial learning rate of 0.1. The warm-up strategy is also adopted. In the second stage, the learning rate is set to 0.01. The total number of epochs is 60, where we train the model for 40 epochs in the first stage and then fine-tune it using the remaining 20 epochs. The parameter $\lambda$ in Eq.\ref{eq:DOTLA} of the DOTLA module is set to 25, empirically. The value of $k$ in Eq.\ref{eq:k}, which denotes the number of neighbors, is set to 10. The value of $\gamma$ in Eq.\ref{gamma} which controls the degree of label refinement and the threshold parameter $\tau$ in Eq.\ref{eq:tau} is set to 0.25 and 1.0, respectively. The trade-off parameter $\alpha$ in Eq.\ref{alpha} is set to 0.3.

\begin{table*}
\small
\caption{Comparison with the state-of-the-art methods on RegDB dataset. It contains supervised VI-ReID (SVI-ReID), unsupervised single-modality ReID (USL-ReID), and unsupervised VI-ReID (USL-VI-ReID) methods. Rank-k accuracy(\%), mAP (\%) and mINP (\%) are reported.}
\label{table 2}

  \centering
  \vspace{-0.3cm}

	\begin{tabular}{c | c c | c c c c c | c c c c c}
		\toprule

		&\multicolumn{2}{c|}{RegDB Settings} & \multicolumn{5}{c|}{Visible to Infrared} & \multicolumn{5}{c}{Infrared to Visible} \\

		&Methods&Venue&r1(\%)&r10(\%)&r20(\%) &mAP(\%)&mINP(\%)&r1(\%)&r10(\%)&r20(\%)&mAP(\%)&mINP(\%)\\
		\hline 
		\multirow{5}{*}{SVI-ReID} 
            & X-Modal~\cite{li2020infrared} & AAAI-20 & 62.21 & 83.13 & 91.72 & 60.18 &-& -&-&-&-&-\\
		&Hi-CMD~\cite{choi2020hi} &CVPR-20&70.93&86.39&-&66.04&-&-&-&-&-&-\\

		&AGW~\cite{ye2021deep}&TPAMI-21&70.05&86.21&91.55&66.37&50.19&70.49&87.21&91.84&65.90&51.24\\
		&CA~\cite{ye2021channel}&ICCV-21&85.03&95.49&97.54&79.14&65.33&84.75&95.33&97.51&77.82&61.56\\
		&FMCNet~\cite{zhang2022fmcnet}&CVPR-22&89.12&-&-&84.43&-&88.38&-&-&83.86&-\\
		\hline 
		\multirow{6}{*}{USL-ReID}&SPCL~\cite{ge2020self}&NIPS-20&13.59&26.98&34.88&14.86&10.36&11.70&25.53&32.82&13.56&10.09\\
		&MMT~\cite{ge2020mutual}&ICLR-20&25.68&42.23&54.03&26.51&19.56&24.42&41.21&51.89&25.59&18.66\\
		&Cluster-Contrast~\cite{dai2022cluster}&arXiv-21&11.76&24.83&32.84&13.88&9.94&11.14&24.11&32.65&12.99&8.99\\
		&ICE~\cite{chen2021ice}&ICCV-21&12.98&25.87&34.4&15.64&11.91&12.18&25.67&34.9&14.82&10.6\\
            
            &PPLR~\cite{cho2022part}&CVPR-22&8.93&20.87&27.91&11.14&7.89&8.11&20.29&28.79&9.07&5.65\\
           & ISE \cite{zhang2022implicit} & CVPR-22 & 16.12 & 23.30 & 28.93 & 16.99 & 13.24 & 10.83 & 18.64 & 27.09 & 13.66 & 10.71  \\
            \hline 
		\multirow{3}{*}{USL-VI-ReID}&H2H~\cite{liang2021homogeneous}&TIP-21&23.81&45.31&54.00&18.87&-&-&-&-&-&-\\
		&OTLA~\cite{wang2022optimal}&ECCV-22&32.9&-&-&29.7&-&32.1&-&-&28.6&-\\
		&ADCA~\cite{yang2022augmented}&MM-22&\underline{67.2}&\underline{82.02}&\underline{87.44}&\underline{64.05}&\underline{52.67}&\underline{68.48}&\underline{83.21}&\underline{88.00}&\underline{63.81}&\underline{49.62}\\
        \hline 
		&Ours&-&\textbf{85.63}&\textbf{94.08}&\textbf{95.49}&\textbf{76.71}&\textbf{61.58}&\textbf{82.91}&\textbf{92.33}&\textbf{94.90}&\textbf{74.97}&\textbf{58.60}\\
            \bottomrule
	\end{tabular}

\end{table*}
\begin{table*}
\small
\caption{ Ablation studies on SYSU-MM01 and RegDB. "DOTLA" denotes the dual-optimal transport for cross-modality label alignment. "CNCR" means the cross-modality neighbor consistency. "NCLR" represents the neighbor consistency guided label refinement. Rank-k accuracy(\%), mAP (\%) and mINP (\%) are reported.}
  \centering
  \setlength\tabcolsep{3pt}
  \vspace{-0.3cm}
  \label{table Ab1:}
    \begin{tabular}{c|cccc|p{0.8cm}<{\centering}p{0.8cm}<{\centering}p{0.8cm}<{\centering}|p{0.9cm}<{\centering}p{0.9cm}<{\centering}p{0.9cm}<{\centering}|p{0.85cm}<{\centering}p{0.85cm}<{\centering}p{0.85cm}<{\centering}|p{0.85cm}<{\centering}p{0.85cm}<{\centering}p{0.85cm}<{\centering}}
	\toprule
	& \multicolumn{4}{c|}{Components} & \multicolumn{3}{c|}{SYSU-MM01(All-search)}& \multicolumn{3}{c|}{SYSU-MM01(Indoor-search)} & \multicolumn{3}{c|}{RegDB(Visible to Infrared)} & \multicolumn{3}{c}{RegDB(Infrared to Visible)}  \\ 
    Index & Baseline & DOTLA & NCLR & CNCR & r1 & r10 & mAP & r1 & r10 & mAP & r1 & r10 & mAP & r1 & r10 & mAP\\
\hline
1& \checkmark& & & & 33.8 & 75.50 &32.93 & 35.95 &79.63 & 45.24 & 39.76 & 60.97 & 35.29 & 39.22 & 61.36 & 34.52\\
2& \checkmark&\checkmark&&& 44.52 &83.29& 41.66 & 45.54 &86.26 & 54.07 & 77.52 & 89.81 & 68.39 & 77.04 & 87.96 & 66.75\\

3&\checkmark&\checkmark&\checkmark&& 45.97 &84.83 & 43.03 & 48.08 & 88.67 & 56.46 & 77.62 & 90.53 & 69.09 & 78.62 & 89.15 & 72.82\\

4&\checkmark&\checkmark&&\checkmark&45.16&85.66&43.03&48.75&89.89&57.38&79.27&91.84&71.91&79.27&89.90&70.08\\

5&\checkmark&\checkmark&\checkmark&\checkmark&\textbf{50.36}&\textbf{89.02}&\textbf{47.36}&\textbf{53.47}&\textbf{92.24}&\textbf{61.73}&\textbf{85.63}&\textbf{94.08}&\textbf{76.71}&\textbf{82.91}&\textbf{92.33}&\textbf{74.97}\\

\bottomrule
	\end{tabular}

\end{table*}

\subsection{Comparison with State-of-the-art Methods}
We compare our approach with state-of-the-art supervised VI-ReID (SVI-ReID) and unsupervised VI-ReID (USL-VI-ReID) to demonstrate its
effectiveness, as shown in Table \ref{table 1} and Table \ref{table 2}. 
Due to the lack of methods for USL-VI-ReID, we add several unsupervised single-modality ReID (USL-ReID) methods to our comparison.\\
\textbf{Comparison with USL-VI-ReID methods.} Our proposed method is compared with existing USL-VI-ReID works, including H2H~\cite{liang2021homogeneous}, OTLA~\cite{wang2022optimal}, and ADCA~\cite{yang2022augmented}. 
The proposed method achieves 47.36\% mAP and 50.36\% Rank-1 accuracy on SYSU-MM01 under the all-search mode, surpassing the state-of-the-art ADCA~\cite{yang2022augmented} by a large margin of 4.63\% mAP and 4.85\% Rank-1 accuracy. While for RegDB, our method improves the mAP and Rank-1 accuracy by a large margin of 12.66\% and 18.43\%.
The results illustrate the effectiveness of our method, in its ability to 
build cross-modality relationships and address the noise in cross-modality label assignment.\\
\textbf{Comparison with USL-ReID methods.}
The results reveal that USL-ReID methods perform poorly on cross-modality datasets, because they are not designed to reduce the modality discrepancy. 
Our method surpasses the best USL-ReID method Cluster-Contrast~\cite{dai2022cluster} by a large margin of 25.36\% mAP on the SYSU-MM01 dataset and about 62.83\% mAP on the RegDB dataset. The results illustrate the importance of bridging the modality gap in the USL-VI-ReID task.\\
\textbf{Comparison with SVI-ReID methods}.
Compared with existing SVI-ReID methods, 
our method outperforms Hi-CMD~\cite{choi2020hi} on the SYSU-MM01 dataset and surpasses many SVI-ReID methods on the RegDB dataset, including X-Modal~\cite{li2020infrared}, Hi-CMD~\cite{choi2020hi}, and AGW~\cite{ye2021deep}. The results show that USL-VI-ReID methods have the potential to achieve performances close to SVI-ReID methods.
However, we still have a significant gap with the SOTA SVI-ReID methods.

\begin{table*}
\small
\caption{ The exploration of influence for each branch on SYSU-MM01 and RegDB datasets. Rank-k accuracy(\%) and mAP (\%) are reported.}
  \centering
  \setlength\tabcolsep{5pt}
  \vspace{-0.3cm}
 \label{table path1:}
	\begin{tabular}{p{3.95cm}<{\centering}|p{0.7cm}<{\centering}p{0.7cm}<{\centering}p{0.7cm}<{\centering}|p{0.7cm}<{\centering}p{0.7cm}<{\centering}p{0.7cm}<{\centering}|p{0.7cm}<{\centering}p{0.7cm}<{\centering}p{0.7cm}<{\centering}|p{0.7cm}<{\centering}p{0.7cm}<{\centering}p{0.7cm}<{\centering}}
	\toprule
	Type of path & \multicolumn{3}{c|}{SYSU-MM01(All-search)}& \multicolumn{3}{c|}{SYSU-MM01(Indoor-search)} & \multicolumn{3}{c|}{RegDB(Visible to Infrared)} & \multicolumn{3}{c}{RegDB(Infrared to Visible)}  \\ 
    & r1 & r10 & mAP & r1 & r10 & mAP & r1 & r10 & mAP & r1 & r10 & mAP\\
\hline
Visible-ReID branch & 40.31 & 81.99 & 39.29 & 45.21 & 87.53 & 54.86 & 79.76 & 90.78 & 70.78 & 75.58 & 88.06 & 68.56\\
Infrared-ReID branch & 47.31 & 86.94 & 44.99 & 50.94 & 90.14 & 58.96 & 70.1 & 85.44 & 64.37 & 75.24 & 86.99 & 68.68\\
Collaborative learning &\textbf{50.36}&\textbf{89.02}&\textbf{47.36}&\textbf{53.47}&\textbf{92.24}&\textbf{61.73}&\textbf{85.63}&\textbf{94.08} &\textbf{76.71}  &\textbf{82.91}& \textbf{92.33}&\textbf{74.97}\\

\bottomrule	
\end{tabular}
\end{table*}

\begin{figure}
  \centering
  \includegraphics[width=8.5cm]{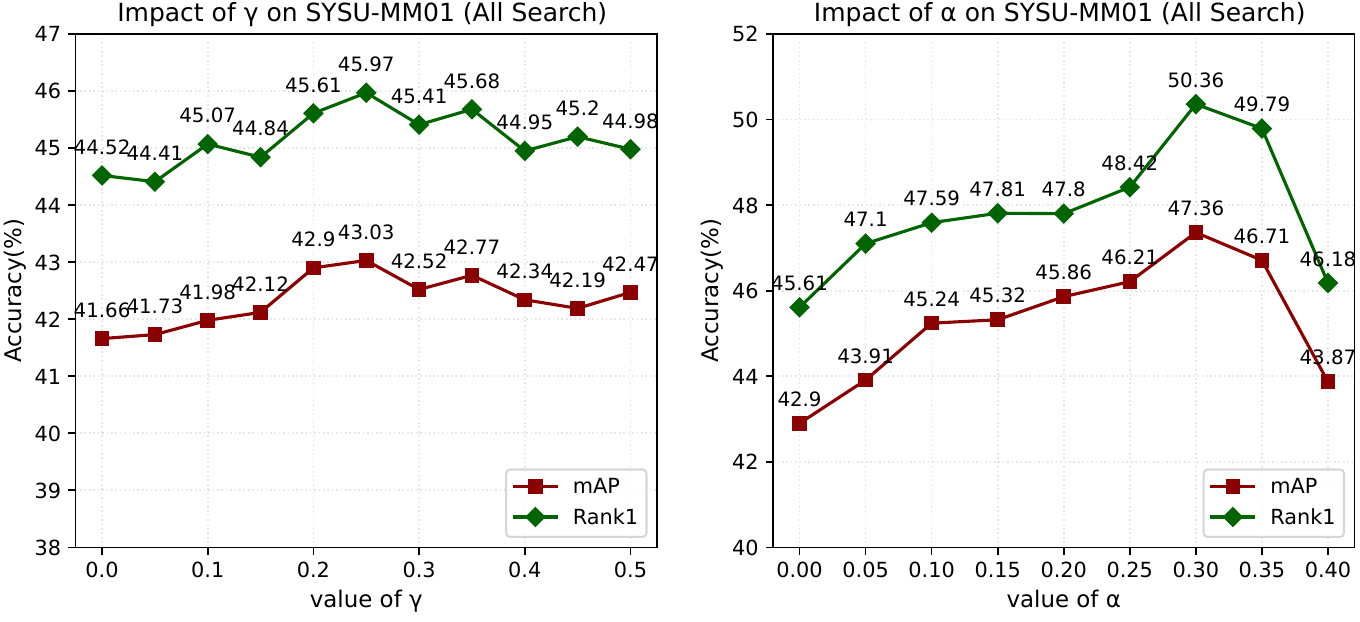}\\
  \vspace{-2mm}
  \caption{The hyper-parameter analysis of $\gamma$ and $\alpha$ on SYSU-MM01 dataset}\label{parameter-analysis}
  \vspace{-3mm}
\end{figure}

\begin{table}
\small
\caption{The validation of robustness on RegDB dataset. "w/o pretraining" means that we train the model without pretraining on labeled Market-1501 dataset. Rank-k accuracy(\%) and mAP (\%) are reported.}
\label{table 3}

  \centering

	\begin{tabular}{c | c c| c c }
		\toprule

		Backbone & \multicolumn{2}{c|}{Visible to Infrared} & \multicolumn{2}{c}{Infrared to Visible} \\

		&r1(\%) &mAP(\%)&r1(\%)&mAP(\%)\\
            \hline
            ResNet50(w/o pretraining) & 74.66 & 68.70 & 73.88 & 67.64 \\
		\hline
            AGW(w/o pretraining) & 79.37 & 72.91 & 78.01& 71.15 \\

        \hline 
		AGW(w/ pretraining)&\textbf{85.63}&\textbf{76.71}&\textbf{82.91}&\textbf{74.97}\\
            \bottomrule
	\end{tabular}

\end{table}

\subsection{Ablation Study}
To evaluate the effectiveness of each component in our proposed method, we conduct ablation studies on the SYSU-MM01 and RegDB datasets, as shown in Table \ref{table Ab1:}.\\
\textbf{Baseline} denotes that we train the model with Eq.\ref{equ:baseline}, which can learn modality-shared representations to some extent by projecting the visible and infrared modalities into a common space. \\
\textbf{Effectiveness of DOTLA.} 
The efficacy of the DOTLA module is revealed when comparing Index 1 and Index 2 in Table \ref{table Ab1:}.
The DOTLA significantly improves 8.73\% mAP on the SYSU-MM01 dataset. 
DOTLA aligns the labels in a collaborative manner to build relationships between cross-modality instances, which 
can alleviate the noisy correspondences.\\
\textbf{Effectiveness of NCLR.} The results of Index 2 and Index 3 show the efficacy of NCLR, which improves the performance by 1.37\% mAP on SYSU-MM01 dataset.
The main improvement is from the data associations refined by the cross-modality neighbors.\\
\textbf{Effectiveness of CNCR.} Comparison between Index 3 and Index 5 shows that the CNCR improves 4.33\% mAP on the SYSU-MM01 dataset.
By encouraging the predictions of images from two modalities to be similar if they are close in the embedding space, the model can be more robust to noisy labels.

\begin{figure}
  \centering
  \includegraphics[width=8.5cm]{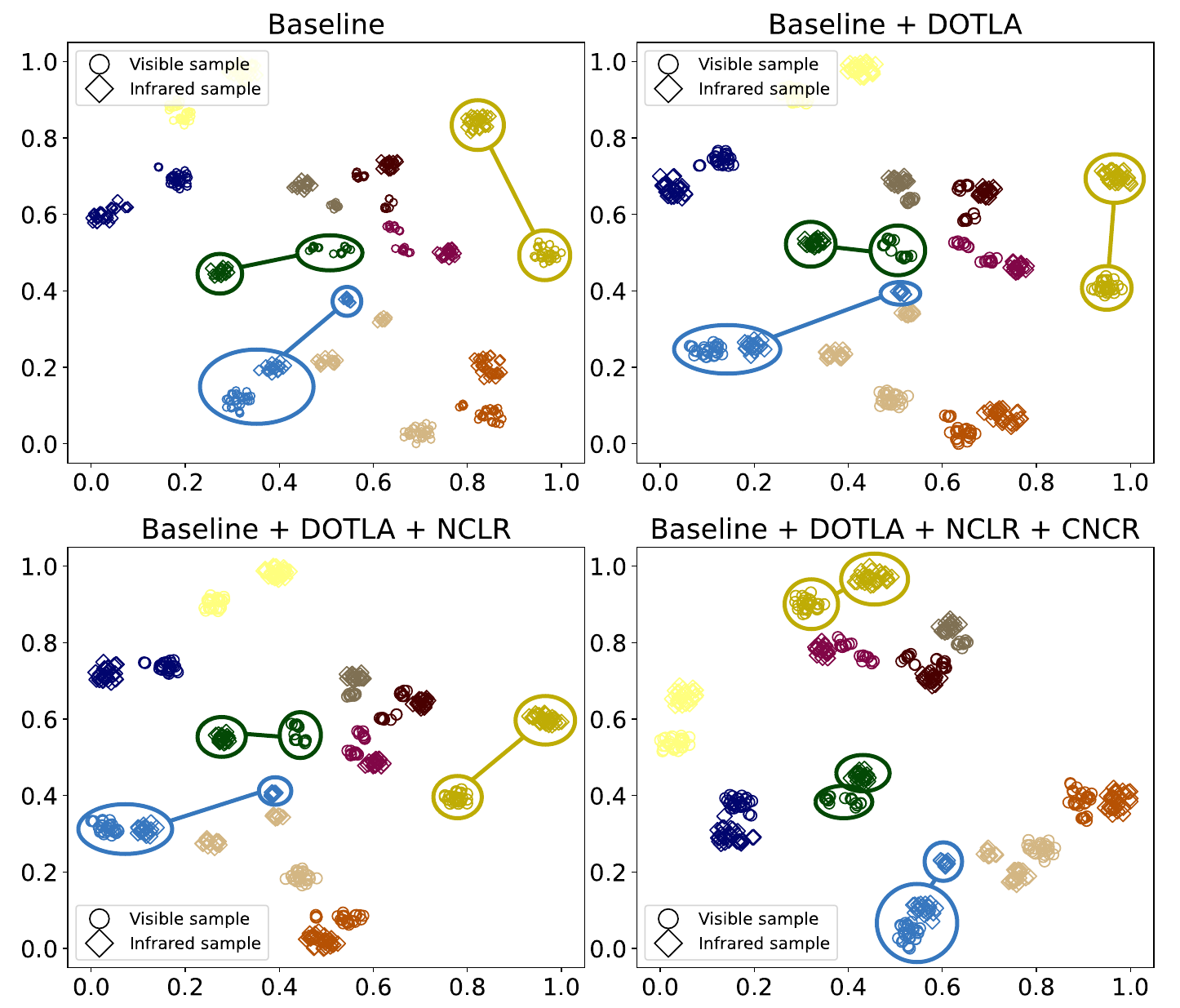}\\
  \caption{The t-SNE\cite{van2008visualizing} visualization of learned features for 10 randomly selected identities. 
  Different colors represent different ground-truth identities. "$\mathbf{\circ}$"  denotes the samples from visible  modality while "$\mathbf{\Diamond}$" from infrared modality.  
  }\label{tsne}
  \vspace{-2mm}
\end{figure}

\begin{figure}
  \centering
  \includegraphics[width=8.5cm]{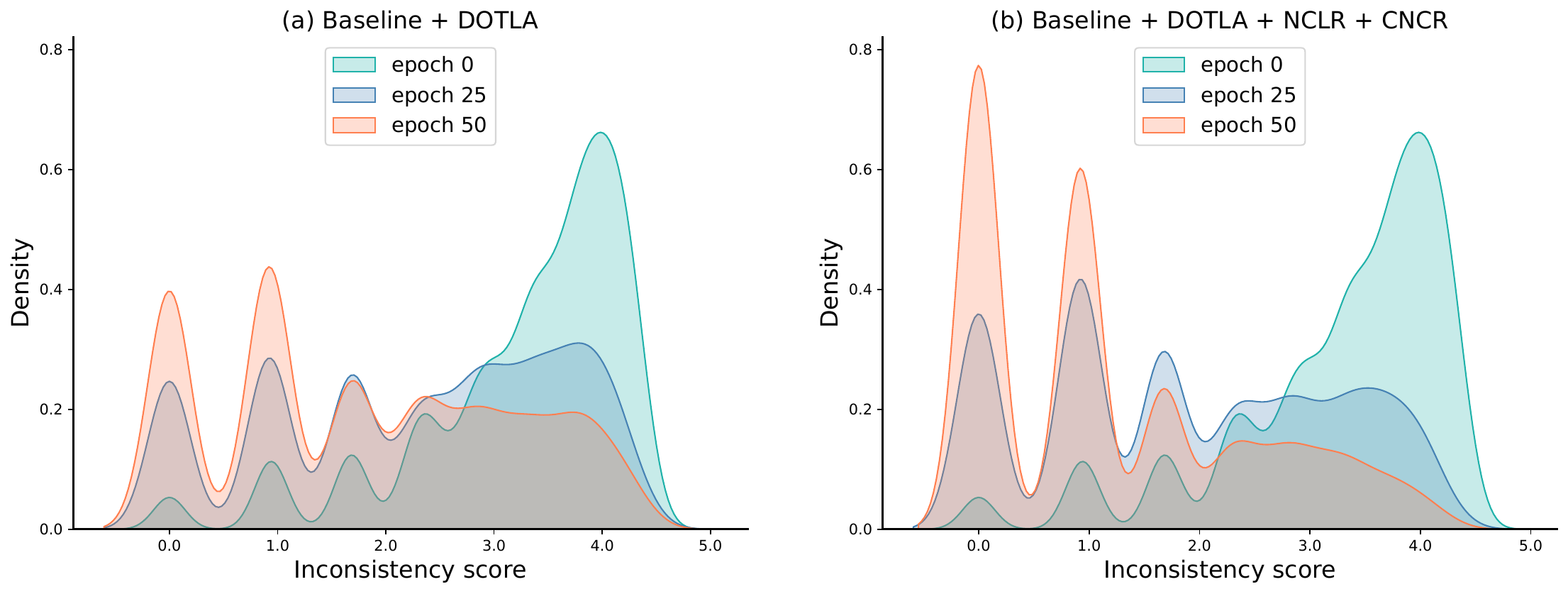}\\
  \caption{The distribution variations of the inconsistency scores 
  of the infrared samples from RegDB dataset 
  during model training. We conduct comparison on (a) baseline+DOTLA and (b) Baseline+DOTLA+NCLR+CNCR.}\label{inconsistency-scores}
  \vspace{-2mm}
\end{figure}

\subsection{Further Analysis}
\textbf{Hyper-parameter Analysis}. We explore the influence of two hyper-parameters in our method: the degree parameter $\gamma$ for label refinement in Eq. \ref{gamma} and the trade-off parameter $\alpha$ for CNCR in Eq. \ref{alpha}. We tune the value of each parameter while keeping the other fixed, and the results are shown in Figure \ref{parameter-analysis}. 
It illustrates that too large $\gamma$ and too large $\alpha$ both cause performance degradation. 
Based on these experiments, we set $\gamma = 0.25$ and $\alpha = 0.3$.\\
\textbf{Collaborative Learning}. 
In this section, we analyze the impact of each branch in our collaborative learning framework, and the results are shown in Table \ref{table path1:}. Visible-ReID branch in Table \ref{table path1:} denotes that we keep only the visible branch during training, and vice versa.
From Table \ref{table path1:}, it can be found that only keeping the visible branch and infrared branch leads to a considerable drop of 8.07\%mAP and 2.37\%mAP on SYSU-MM01 dataset, respectively. The results demonstrate that the collaborative learning manner is necessary for handling noisy label assignments. \\
\textbf{T-SNE Visualization.} To illustrate the impact of our method on learning modality-invariant features, we randomly select 10 identities from SYSU-MM01, and then visualize the features using t-SNE, are shown in Figure \ref{tsne}. As we gradually add designed modules to the baseline, samples from the same identity become closer.\\
\textbf{Distribution Variations of Inconsistency Scores.} We visualize distributions of the inconsistency scores of infrared samples from RegDB dataset in Figure~\ref{inconsistency-scores}. Clearly, we can see that more and more examples become consistent with their neighbours (lower $s(\mathbf{x}_i^{r})$) as training progresses. By comparing methods Baseline+DOTLA and Baseline+DOTLA+NCLR+CNCR, we can see that the modules NCLR+CNCR make better use of the neighbour consistency constraint, and thus help to improve model performance.\\
\textbf{Verifying robustness.} We conduct experiments on RegDB dataset without pretraining on labeled Market-1501 dataset. 
AGW and ResNet50 are used as backbones. The results in Table \ref{table 3} illustrate the robustness of our method.   


\section{Conclusion}
This paper proposes a two-stream collaborative learning framework followed by the label refinement and regularization under the guidance of cross-modality neighbor consistency for USL-VI-ReID task. We design a dual optimal transport label assignment module to assign labels from one modality to its counterpart modality, which formulates a mutual reinforcement and efficient solution to cross-modality label association.
Additionally, we further propose a cross-modality neighbor consistency guided label refinement and regularization module, to eliminate the negative effects brought by the inaccurate supervised signals. In the future, we will extend the proposed method to other cross-modality applications. 

\section{Acknowledgments}
This work was supported in part by the National Natural Science Foundation of China under Grant 62176198, U22A2096 and 62036007, in part by the Technology Innovation Leading Program of Shaanxi under Grant 20220FY01-15, in part by Open Research Projects Zhejiang Lab under Grant 2021KG0AB01.  


\bibliographystyle{ACM-Reference-Format}
\balance
\bibliography{sample-base}


\end{document}